\documentclass{article} 
\usepackage{iclr2017_conference,times}
\usepackage{hyperref}
\usepackage{url}
\usepackage{amsmath}
\usepackage{algorithm}
\usepackage{algpseudocode}
\usepackage{graphicx}

\title{Generative Adversarial Networks as Variational Training of Energy Based Models}

\author{Shuangfei Zhai \\
  Binghamton University\\
  Vestal, NY 13902, USA \\
  \texttt{szhai2@binghamton.edu}\\
  \And 
  Yu Cheng \\
  IBM T.J. Watson Research Center \\
  Yorktown Heights, NY 10598, USA \\
  \texttt{chengyu@us.ibm.com} \\
  \And 
  Rogerio Feris \\
  IBM T.J. Watson Research Center \\
  Yorktown Heights, NY 10598, USA \\
  \texttt{rsferis@us.ibm.com} \\
  \And
  Zhongfei (Mark) Zhang \\
  Binghamton University\\
  Vestal, NY 13902, USA \\
  \texttt{zhongfei@cs.binghamton.edu} \\
}

\begin{document}

\maketitle

\begin{abstract}
In this paper, we study deep generative models for effective unsupervised learning. We propose VGAN, which works by minimizing a variational lower bound of the negative log likelihood (NLL) of an energy based model (EBM), where the model density $p(\mathbf{x})$ is approximated by a variational distribution $q(\mathbf{x})$ that is easy to sample from. The training of VGAN takes a two step procedure: given $p(\mathbf{x})$, $q(\mathbf{x})$ is updated to maximize the lower bound; $p(\mathbf{x})$ is then updated one step with samples drawn from $q(\mathbf{x})$ to decrease the lower bound. VGAN is inspired by the generative adversarial networks (GANs), where $p(\mathbf{x})$ corresponds to the discriminator and $q(\mathbf{x})$ corresponds to the generator, but with several notable differences. We hence name our model variational GANs (VGANs). VGAN provides a practical solution to training deep EBMs in high dimensional space, by eliminating the need of MCMC sampling. From this view, we are also able to identify causes to the difficulty of training GANs and propose viable solutions. \footnote{Experimental code is available at https://github.com/Shuangfei/vgan}
\end{abstract}

\section{Introduction}
Unsupervised learning is a long standing challenge of machine learning and deep learning. One major difficulty of effective unsupervised learning lies in the lack of an accurate distance metric. Euclidean distance has been widely adopted as the default metric from shallow methods, such as K-means and Gaussian mixture models, to deep models such as autoencoder variants (e.g., \cite{dae}). From a probabilistic point of view, the use of Euclidean distance assumes Gaussian distributions (or mixtures thereof) in the input space, which is a strong assumption and is often times inaccurate for high dimensional data (such as images). Generative adversarial networks (GANs) \citealt{gan} are a particularly interesting approach as it does not assume the data distribution to take any specific form, which therefore eliminates the need of a predefined distance metric of samples. GANs work as a mini-max two player game, where a generator $G(\mathbf{z})$ is trained to generate samples that can fool the best discriminator D. When both G and D are formulated as deep convolutional networks, it is shown that the generator can learn to generate surprisingly realistic looking images \cite{dcgan}. Energy-based models (EBMs) \cite{ebm} are another powerful family of unsupervised learning models. Similarly to GANs, EBMs make minimal assumptions about the data distribution, as it directly parameterizes the negative long density of data $E(\mathbf{x}) = -\log p(\mathbf{x})$ as a deterministic function of $\mathbf{x}$. It is clear that by properly choosing the capacity of $E(\mathbf{x})$, an EBM can be trained to approximate an arbitrary density function perfectly well.

In this paper, we propose VGAN, which bridges GANs and EBMs and combines the benefits from both worlds. In particular, we show that the mini-max game of GANs is approximately equivalent to minimizing a variational lower bound of the negative log likelihood (NLL) of an EBM. To be more concrete, the energy $E(\mathbf{x})$ corresponds to $-\log D(\mathbf{x})$, and the generator $G(\mathbf{z})$ defines a parameterzied sampler from the model distribution defined by $p(\mathbf{x}) = \frac{e^{-E(\mathbf{x})}}{\int_x{e^{-E(\mathbf{x})}}dx}$. From this view, GANs provide a viable solution for the maximum likelihood estimation of EBMs, which is known to be challenging due to the difficulty of evaluating the partition function which integrates over the input space. We discuss the important design choices of the energy functions in order to make VGAN numerically stable, and propose a novel energy formulation that is bounded and explicitly multi-modal. Moreover, from the EBM point of view, we are also able to identify the reasons that make GANs unstable to train, due to the missing of an entropy term of the generator distribution, which causes the generator to collapse to a single or few local minima of the energy landscape. As a solution, we propose to parameterize the generator as a transition distribution (that is, $p_z(\mathbf{\tilde{x}}|\mathbf{x})$ instead of $p_z(\mathbf{x})$), in analogy to the one used in Gibbs sampling procedure. We show that this variant corresponds to a variational version of contrastive divergence \cite{cd}, and circumvents the need of directly approximating the cumbersome entropy term. In our experiments on MNIST, CIFAR10, and SVHN, we show that we are able to learn generators that generate sharp and diversified images. Moreover, the learned transition distributions are able to effectively capture the data manifold by consecutively sampling realistic looking samples  starting from testing images. Finally, as a quantitative evaluation of the learned model, we use the transition distribution as data augmentation, from which we are able to show consistent gains of classification accuracy with few training labels on MNIST and SVHN.

\section{Generative Adversarial Networks}
Generative adversarial networks \cite{gan} work by solving the following mini-max game:
\begin{equation}
\label{eq:gan}
\max_G \min_D \mathrm{E}_{\mathbf{x} \sim p_{data}(\mathbf{x})}[-\log D(\mathbf{x})] - \mathrm{E}_{\mathbf{z} \sim p_{\mathbf{z}}(\mathbf{\mathbf{z}})}[\log(1 - D(G(\mathbf{z})))],
\end{equation}
where $p_{data}(\mathbf{x})$ is the data distribution; $D(\mathbf{x})$ is the discriminator that takes as input a sample and outputs a scalar between $[0,1]$; $G(\mathbf{z})$ is the generator that maps a sample $\mathbf{z} \in R^d$ drawn from a simple distribution $p(\mathbf{z})$ to the input space. Typically both $D$ and $G$ are parameterized as deep neural networks. Equation \ref{eq:gan} suggests a training procedure consisting of two loops: in the inner loop D is trained till convergence given G, and in the outer loop G is updated one step given D (note that in \cite{gan}, the authors propose to maximize $\log(D(G(\mathbf{z})))$ instead of $-\log(1 - D(G(\mathbf{z})))$ in the outer loop). As the two-player, mini-max game reaches the Nash equilibrium, G defines an implicit distribution $p_g(\mathbf{x})$ that recovers the data distribution, i.e., $p_g(\mathbf{x}) = p_{data}(\mathbf{x})$.

\section{Variational Training of Deep-EBMs}
An EBM formulates a density function as:
\begin{equation}
\label{eq:ebm}
p(\mathbf{x}) = \frac{e^{-E(\mathbf{x})}}{\int_{\mathbf{x}}{e^{-E(\mathbf{x})}dx}},
\end{equation}
where $E(\mathbf{x})$ is defined as the energy of input $\mathbf{x}$. One particularly powerful case is deep energy based models (deep-EBMs) \cite{debm,dsebm}, where $E(\mathbf{x})$ is directly parameterized as the output of a deep deterministic neural network. An obvious way to train an EBM is to minimize the negative log likelihood (NLL):
\begin{equation}
\label{eq:nll}
J(E) = \mathrm{E}_{\mathbf{x} \sim p_{data}(\mathbf{x})} [E(\mathbf{x})] + \log [\int_{\mathbf{x}}{e^{-E(\mathbf{x})}d\mathbf{x}}].
\end{equation}
Directly minimizing $J(E)$ is difficult due to the integration term over the input space. As a remedy, one can rewrite Equation \ref{eq:nll} as follows:
\begin{equation}
\begin{split}
J(E) &= \mathrm{E}_{\mathbf{x} \sim p_{data}(\mathbf{x})} [E(\mathbf{x})] + \log [\int_{\mathbf{x}}{q(\mathbf{x})\frac{ e^{-E(\mathbf{x})}}{q(\mathbf{x})}d\mathbf{x}}] \\
&= \mathrm{E}_{\mathbf{x} \sim p_{data}(\mathbf{x})} [E(\mathbf{x})] + \log [\mathrm{E}_{\mathbf{x} \sim q(\mathbf{x})}[\frac{ e^{-E(\mathbf{x})}}{q(\mathbf{x})}]] \\
&\geq \mathrm{E}_{\mathbf{x} \sim p_{data}(\mathbf{x})} [E(\mathbf{x})] + \mathrm{E}_{\mathbf{x} \sim q(\mathbf{x})}[{\log \frac{ e^{-E(\mathbf{x})}}{q(\mathbf{x})}]} \\
&= \mathrm{E}_{\mathbf{x} \sim p_{data}(\mathbf{x})} [E(\mathbf{x})] - \mathrm{E}_{\mathbf{x} \sim q(\mathbf{x})}[E(\mathbf{x})] + H(q),
\end{split}
\label{eq:variational_nll}
\end{equation}
where $q(\mathbf{x})$ is an arbitrary distribution which we call the \textit{variational distribution} with $H(q)$ denoting its entropy. Equation \ref{eq:variational_nll} is a natural application of the Jensen's inequality, and it gives a variational lower bound of the NLL given an $q(\mathbf{x})$. The lower bound is tight when $\frac{ e^{-E(\mathbf{x})}}{q(\mathbf{x})}$ is a constant independent of $\mathbf{x}$, i.e., $q(\mathbf{x}) \propto E(\mathbf{x})$, which implies that $q(\mathbf{x}) = p(\mathbf{x})$. This suggests an optimization procedure as follows:
\begin{equation}
\label{eq:minmax}
\min_{E}\max_{q} \mathrm{E}_{\mathbf{x} \sim p_{data}(\mathbf{x})} [E(\mathbf{x})] - \mathrm{E}_{\mathbf{x} \sim q(\mathbf{x})}[E(\mathbf{x})] + H(q),
\end{equation}
where in the inner loop, given the energy model $E(\mathbf{x})$, the variational lower bound is maximized w.r.t. q; the energy model then is updated one step to decrease the NLL with the optimal $q$.

\section{Variational Generative Adversarial Networks}
\label{sec:vgan}
In practice, $q(\mathbf{x})$ can be chosen as a distribution that is easy to sample from and differentiable; the inner loop can be achieved by simple stochastic gradient descent. It turns out that the generator used in GANs exactly satisfies such requirements, which directly connects GANs to EBMs. To see this, replace $q(\mathbf{x})$ with the generator distribution $p_g(\mathbf{x})$ (that is the implicit data distribution produced by $\mathbf{x}=G(\mathbf{z}), \mathbf{z} \sim p_z(\mathbf{z})$), then Equation \ref{eq:minmax} turns into:
\begin{equation}
\label{eq:vgan}
\min_{E}\max_{G} \mathrm{E}_{\mathbf{x} \sim p_{data}(\mathbf{x})} [E(\mathbf{x})] - \mathrm{E}_{\mathbf{x} \sim p_{g}(\mathbf{x})}[E(\mathbf{x})] + H(p_g).
\end{equation}
If we further let $E(\mathbf{x}) = -\log D(\mathbf{x})$, this becomes:
\begin{equation}
\label{eq:ebm_as_gan}
\min_{D}\max_{G} \mathrm{E}_{\mathbf{x} \sim p_{data}(\mathbf{x})} [-\log D(\mathbf{x})] - \mathrm{E}_{\mathbf{z} \sim p_{\mathbf{z}}(\mathbf{z})}[-\log D(G(\mathbf{z}))] + H(p_g).
\end{equation}
One can now immediately recognize the resemblance of Equation \ref{eq:ebm_as_gan} to Equation \ref{eq:gan}. Both of them take the form as a mini-max optimization problem, where D is trained to increase $D(\mathbf{x})$ for $\mathbf{x} \sim p_{data}(\mathbf{x})$ and decrease $D(\mathbf{x})$ for $\mathbf{x} \sim p_g(\mathbf{x})$, while G is trained to increase $D(\mathbf{x})$ for $\mathbf{x} \sim p_g(\mathbf{x})$. In other words, GAN behaves similarly to variational training of an EBM, where the variational distribution $q(\mathbf{x})$ takes the specific form of $p_g({\mathbf{x}})$ which is easy to sample from. In light of this connection, we call the family of the models defined by Equation \ref{eq:vgan} as the variational generative adversarial networks (VGANs). The nice property of VGANs over the traditional EBM training strategies is that it simplifies the sampling procedure by defining a differentiable, deterministic mapping from a simple distribution (e.g., uniform distribution) to the input space.

Compared with GANs, VGANs differ in four aspects:

1. \textbf{The order of minimization and maximization}. GANs optimize D till convergence given G, while VGANs optimize G till convergence given D. The outcome of a GAN is then a generator that can fool the best discriminator possible, while the outcome of a VGAN is an energy model parameterized by D, and a variational distribution G that can sample from the exact distribution defined by D. Also note that with the optimization procedure of GANs, there is no guarantee that D defines a viable EBM, as the variational lower bound can be arbitrarily low due to the swapping of the min and max loops.

2. \textbf{The parameterization of energy}. GANs use a specific parameterization of energy as $E(\mathbf{x}) = -\log D(\mathbf{x})$. The energy parameterization of GANs is lower bounded by $0$. This differs from that of an RBM with binary visible hidden units, one of the most popular EBMs. An RBM has free energy $E(\mathbf{x}) = -\mathbf{b_v}^T\mathbf{x} - \sum_{j=1}^K{\log(1 + e^{\mathbf{W}_j^T\mathbf{x} + \mathbf{b}_{h,j}})}$, which is unbounded. As the goal of training an EBM is to minimize the energy of training data, this difference is significant in practice. To see this, note that the optimum in Equation \ref{eq:minmax} is invariant w.r.t. an affine transformation of the energy; that is, let $E^*(\mathbf{x})$ be an optimal solution to Equation \ref{eq:minmax}; then $\tilde{E}(\mathbf{x}) = a E^*(\mathbf{x}) + b$ is also an optimal solution for $a \in R^+, b \in R$. This property makes unbounded energy inappropriate to use for VGANs, as it often causes the scale of energy to explode. Even worse, the energy parameterization as that of RBM's has stronger gradients as the energy decreases, and this essentially encourages the energy of both training samples and generated samples to grow to negative infinity.

3. \textbf{The optimal energy assignment}. A related problem to the energy parameterization is that, when optimizing $D$, the term subject to expectation under $p(\mathbf{z})$ of GANs is $ \log(1- D(G(\mathbf{z})))$, whereas VGANs use $-\log D(G(\mathbf{z}))$. While both have the same direction of gradients w.r.t. to $D(\mathbf{x})$ and $D(G(\mathbf{z}))$ (increasing the former and decreasing the latter), and the optimal solution to both models is when $p_{data}(\mathbf{x}) = p_g(\mathbf{x})$, they differ in the optimal $D$. The optimal $D$ for GAN is fixed as $D(\mathbf{x}) = 0.5$. 

4. \textbf{The entropy term of the generator distribution $H(p_g)$}. Last but not the least, GANs do not include the entropy term while optimizing $G$. In VGANs, including the entropy term guarantees that $p_g(\mathbf{x})$ recovers the density encoded by $D$, and that the variational lower bound is tightened as such in the inner loop. Without the  entropy term, $G$ can be easily but misleadingly optimized by collapsing into one of the few local minima of the energy landscape. In fact, this accounts for most of the failures of training GANs, as pointed out in the GAN related literature \cite{dcgan,improvedgan,kim2016deep,zhao2016energy}. Of course, an immediate challenge that one needs to solve is the approximation of $H(p_g)$. This amounts to a well known problem of differentiable entropy approximation (see \cite{entropy}, for example). The fact that the approximation needs not only to be accurate, but also to be easily optimized w.r.t. G makes it even more intimidating and cumbersome.

\section{Bounded Multi-modal Energy}
\label{sec:energy}
The first strategy we attempt to stabilize the generator is by designing a well behaved energy such that the generator can be easily optimized. We start by noticing that the energy of the form $-\log D(\mathbf{x})$ is inherently uni-modal. To see this, let $D(\mathbf{x}) = \sigma(\mathbf{w}^T\phi(\mathbf{x}) + \mathbf{b})$, where $\sigma$ is the sigmoid function and $\phi(\mathbf{x})$ denotes a feature mapping of $\mathbf{x}$ encoded by a deep neural network. Then in order to maximize $D(\mathbf{x})$ such as to minimize the energy, all the samples $\mathbf{x}$ are thus encouraged to be projected to be proportional to the weight vector $\mathbf{w}$. This is not a problem with the regularization of $H(p_g)$, maximizing which diversifies $\phi(\mathbf{x})$, but without or with a poor approximation of the entropy term may cause the generator to collapse. Consequently, we propose a bounded multi-modal energy formulation as follows:
\begin{equation}
E(\mathbf{x}) = \sum_{j=1}^{K}H(\sigma(\mathbf{W}_j^T\phi(\mathbf{x}) + \mathbf{b}_j)),
\end{equation}
\label{eq:energy}
where $\mathbf{W}_j \in R^d, \mathbf{b}_j \in R$, $\phi(\mathbf{x})$ is the feature mapping; $H(p)$ is slightly overloaded as the entropy of a binomial distribution defined by $p$, i.e., $H(p) = -p\log p - (1-p)\log(1-p)$. This energy formulation can be viewed as an instance of the product of experts model (PoE) \cite{poe}, where each set of parameters $\mathbf{W}_j, \mathbf{b}_j$ defines an expert. The nice property of this energy parameterization is that it is 1) bounded between [0, K], 2) with strong gradients in the high energy area ($\sigma(\cdot)$ close to 0.5) and with vanishing gradients at low energy area ($\sigma(\cdot)$ close to 0 or 1), and 3) multi-modal by design. To see the last point, simply note that $H(p)$ achieves its minimum with $p=0$ and $p=1$. Thus for such a PoE energy with K experts, there exists $2^K$ equally likely local minima by design. With this energy formulation, it is also relatively easy to come up with a reasonable approximation of $H(p_g)$, which is chosen as:
\begin{equation}
\label{eq:entropy}
\tilde{H}(p_g) = \sum_{j=1}^K H(\frac{1}{N}\sum_{i=1}^N\sigma(\mathbf{W}_j^T\phi(G(\mathbf{z})^i))),
\end{equation}
where $\mathbf{x}^i$ denotes the $i$th training example. Although there is no theoretical guarantee that Equation \ref{eq:entropy} recovers the true entropy $H(p_g)$ to any extent, maximizing it serves the same purpose of encouraging the generated samples to be diverse, as $\tilde{H(p_g)}$ reaches its minimum if $G(\mathbf{z})$ collapses to one single point. Moreover, in the outer loop while minimizing the NLL w.r.t. E, we find it helpful to also maximize $\tilde{H}(p_{data}) = \sum_{j=1}^K H(\frac{1}{N}\sum_{i=1}^N\sigma(\mathbf{W}_j^T\phi(\mathbf{x}^i)))$ as well, which acts as a regularizer to $E$ to encourage the average activation of each expert $\sigma_j(\cdot)$ to close to 0.5. The training algorithm of VGAN with the proposed bounded multi-modal energy is summarized in Algorithm \ref{alg:vgan}.
\begin{algorithm}[h]
\caption{The optimization procedure of VGAN}\label{alg:vgan}
\begin{algorithmic}[1]
\small
\For{number of training iterations}
\For{k steps}
\State sample $N$ noise data \{$\mathbf{z}^1$, \dots, $\mathbf{z}^N$\}; update $G$ by one step gradient \textbf{ascent} of
\begin{equation*}
-\frac{1}{N}\sum_{i=1}^N E({G(\mathbf{z}^i)}) + \tilde{H}(p_g)
\end{equation*}
\EndFor
\State sample $N$ training data \{$\mathbf{x}^1$, \dots, $\mathbf{x}^N$\}; sample $N$ noise data \{$\mathbf{z}^1$, \dots, $\mathbf{z}^N$\};
\State update E with one step gradient \textbf{descent} of
\begin{equation*}
\frac{1}{N}\sum_{i=1}^N E({\mathbf{x}^i}) - \frac{1}{N}\sum_{i=1}^N E({G(\mathbf{z}^i)}) - \tilde{H}(p_{data})
\end{equation*}
\EndFor
\end{algorithmic}
\end{algorithm}

\section{Variational Contrastive Divergence with Transition Distributions}
Although it is possible to discourage the generator to collapse into a single output by carefully designing the energy function as described in Section \ref{sec:energy}, there is no good way to monitor the quality of the approximation of the entropy term other than manually observing the generated samples. Also, there is no guarantee that the designed approximation is accurate such that the variational lower bound is tight enough to provide correct gradients for updating the energy parameters. In this section, we propose an additional approach to bypass the need of the cumbersome entropy approximation problem. The idea is that instead of generating samples directly from $p_z(\mathbf{z})$, we define a transition operator $p_g(\mathbf{\tilde{x}}|\mathbf{x})$ conditioned on a training sample $\mathbf{\tilde{x}}$. This corresponds to defining the variational distribution $q(\mathbf{x})$ in Equation \ref{eq:variational_nll} as $q(\mathbf{\tilde{x}}) = \int_{\mathbf{x}} p_{data}(\mathbf{x})p_z(\mathbf{\tilde{x}}|\mathbf{x})d\mathbf{x}$. If we further restrict the transition distribution $p_z(\mathbf{x}|\mathbf{\tilde{x}})$ to be one that is closely centered around $\mathbf{\tilde{x}}$, then the entropy term $H(q)$ can be well approximated by the data entropy $H(p_{data})$, which is a constant. The variational lower bound is thus increased by increasing the energy for $\mathbf{\tilde{x}} \sim p_g(\mathbf{\tilde{x}}|\mathbf{x})$. Of course, this parameterizaton limits the shape of the varaitional distribution, and the variational lower bound might never be tightened especially in the early stage of training when the model distribution differs significantly from the data distribution; this nonetheless can provide meaningful gradients to update the energies. In fact, this sampling procedure is closely related to contrastive divergence (CD) \cite{rbm} (one step CD to be exact), whereas in CD the transition distribution can be easily obtained from specific types of EBMs (e.g., RBM). Our approach, on the other hand, uses a parameterized variational distribution to approximate the true transition distribution; we thus name it \textit{variational contrastive divergence} (VCD).

\begin{figure}
\centering
\includegraphics[scale=0.55]{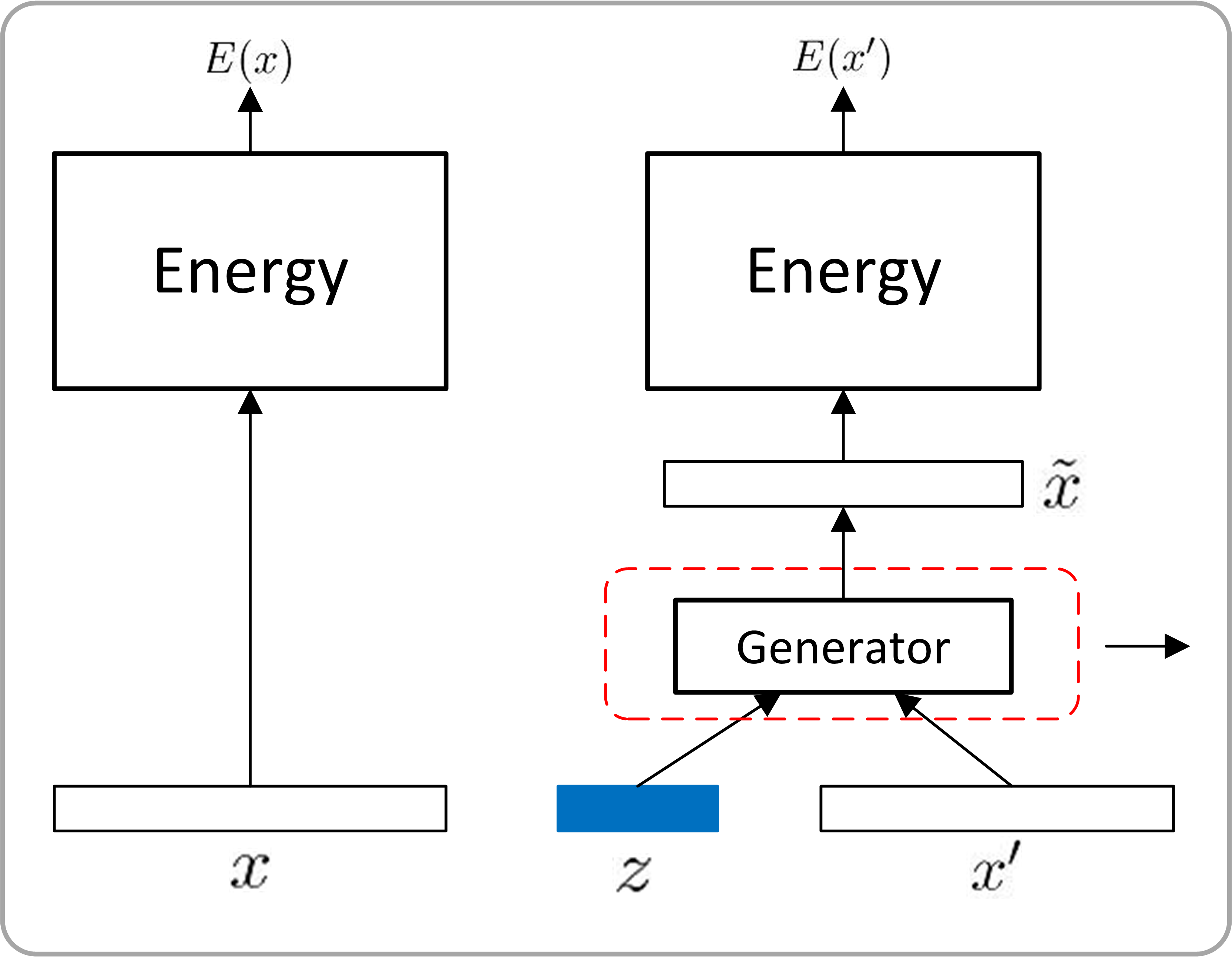}
\hspace{0.1cm}
\includegraphics[scale=0.57]{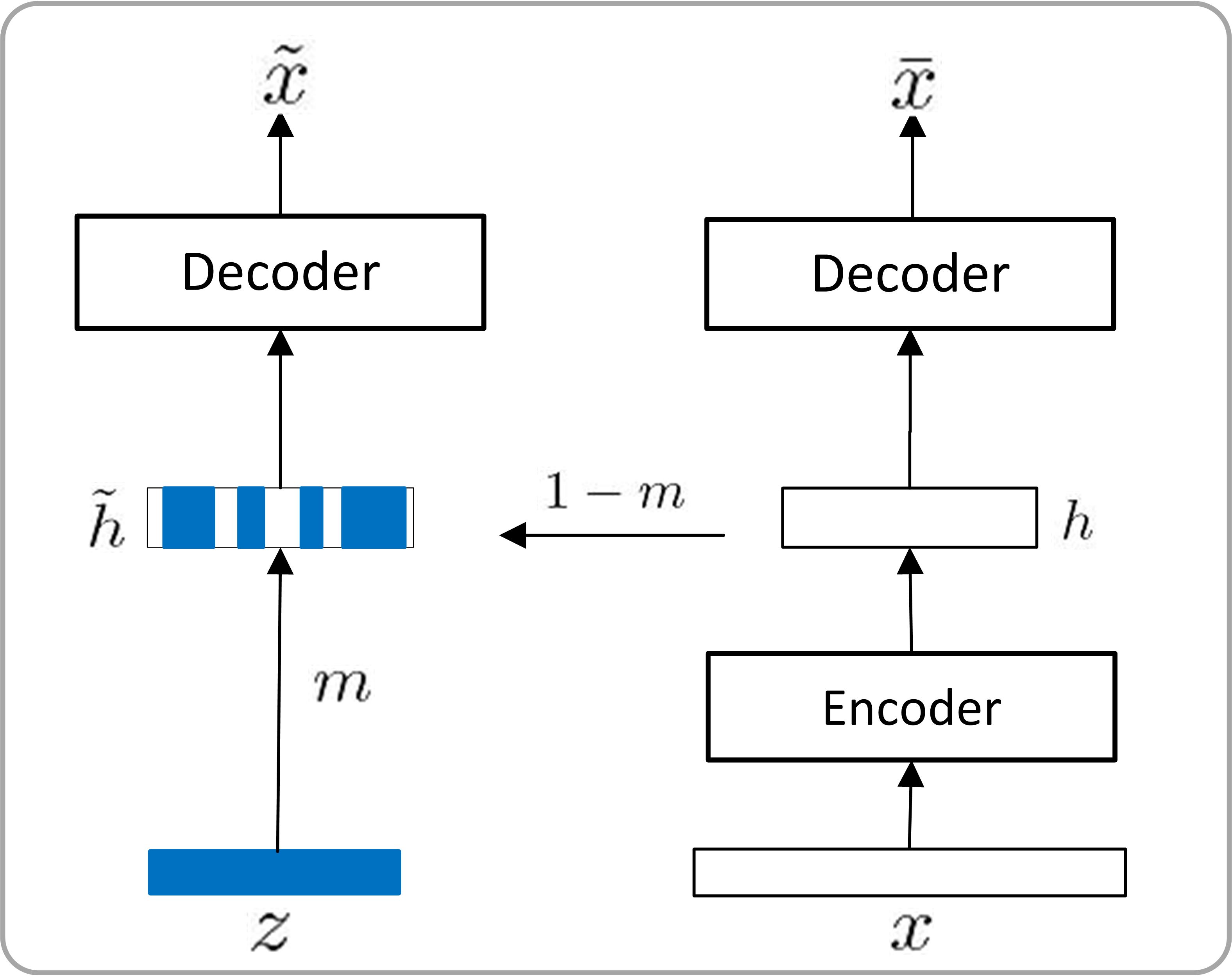}
\caption{Illustration of VGAN with variational contrastive divergence. On the left panel, energies of real data $\mathbf{x}$ and generated data $\mathbf{\tilde{x}}$ are computed, with the generator shown on the right. For the generator on the right panel, each $\mathbf{x} \sim p_{data}(\mathbf{x})$ is passed through an encoder to obtain $\mathbf{h}$, which is then passed through a decoder to achieve a reconstruction $\mathbf{\bar{x}}$. $\mathbf{h}$ is then mixed with a noise vector $\mathbf{z}$ of the same dimensionality by a randomly generated binary mask vector $\mathbf{m}$ to obtain $\mathbf{\tilde{h}}$ following $\mathbf{\tilde{h}} = \mathbf{m} * \mathbf{z} + (\mathbf{1} - \mathbf{m}) * \mathbf{h}$. $\mathbf{\tilde{h}}$ is then passed through the same decoder to obtain the generated sample $\mathbf{\tilde{x}}$.}
\label{fig:vcd}
\end{figure}

The implementation of $p_g(\mathbf{x}|\mathbf{\tilde{x}})$ is illustrated in Figure \ref{fig:vcd}. Let $\mathbf{h} = Encode(\tilde{\mathbf{x}})$ be an encoder that maps an input $\mathbf{\tilde{x}}$ to a bottleneck vector $\mathbf{h} \in R^d$, and let $\bar{\mathbf{x}} = Decode(\mathbf{h})$ be the output of a decoder that maps $\mathbf{h}$ to the input space. A sample from $p_g(\mathbf{\tilde{x}}|\mathbf{x})$ can be drawn as follows: 1) generate a binomial vector $\mathbf{m} \in R^d$ with probability $0.5$; 2) generate a noise vector $\mathbf{z} \sim p_z(\mathbf{z}), \mathbf{z} \in R^d$; 3) produce a new vector $\mathbf{\tilde{h}} = \mathbf{m} * \mathbf{z} + (1 - \mathbf{m}) * \mathbf{h}$; and 4) obtain the generated sample by passing $\mathbf{\tilde{h}}$ to the same decoder $\mathbf{\tilde{x}} = Decode(\mathbf{\tilde{h}})$. The generator then tries to minimize the following objective:
\begin{equation}
\label{eq:ae}
\rho *\mathrm{E}_{\mathbf{x} \sim p_{data}(\mathbf{x}),\mathbf{\tilde{x}} \sim p_g(\mathbf{\tilde{x}}|\mathbf{x})} [E(\mathbf{\tilde{x}})] + (1-\rho)*\mathrm{E}_{\mathbf{x} \sim p_{data}(\mathbf{x})}\|\bar{x} - \mathbf{x}\|^2_2.
\end{equation}
The generator can be considered as a regularized autoencoder, where decreasing the first term of Equation \ref{eq:ae} encourages the EBM to assign low energy to samples generated by $\mathbf{\tilde{h}}$. The choice of the generation formula of $\mathbf{\tilde{h}}$ is also critical. Randomly replacing half of the dimensions of $\mathbf{h}$ with random noise $\mathbf{z}$ makes sure that $\mathbf{\tilde{h}}$ is sufficiently different from $\mathbf{h}$. Otherwise, the autoencoder can easily denoise $\mathbf{\tilde{h}}$ to make $\mathbf{\tilde{x}}$ to collapse back to $\mathbf{x}$, regardless of $\mathbf{z}$. Also, mixing noise in the bottleneck layer of an autoencoder makes the generation process easier, as it is known that with high level features the mixing rate of MCMC sampling is significantly higher than the in the input space \cite{mixing}. In addition, the formulation of our transition operator does not make any Gaussian (or mixture of Gaussian) distribution assumptions, despite the use of the reconstruction error. This is due to the use of a deep decoder, such that the generated sample can be far away from the sample conditioned on, when calculating the Euclidean distance. This conjecture is also supported in our experiments, see Figure \ref{fig:transition}. The training algorithm for VCD is summarized in Table \ref{alg:vcd}. 

\begin{algorithm}[h]
\caption{The optimization procedure of VCD}\label{alg:vcd}
\begin{algorithmic}[1]
\small
\For{number of training iterations}
\For{k steps}
\State sample $N$ training data \{$\mathbf{x}^1$, \dots, $\mathbf{x}^N$\}; sample $N$ noise data \{$\mathbf{z}^1$, \dots, $\mathbf{z}^N$\}; 
\State sample $N$ binary mask vectors; 
\State update $G$ by one step gradient \textbf{ascent} of
\begin{equation*}
-\frac{1}{N}\sum_{i=1}^N E({G(\mathbf{z}^i, \mathbf{m}^i)}) + \tilde{H}(p_g)
\end{equation*}
\EndFor
\State sample $N$ training data \{$\mathbf{x}^1$, \dots, $\mathbf{x}^N$\}; sample $N$ noise data \{$\mathbf{z}^1$, \dots, $\mathbf{z}^N$\}; 
\State sample $N$ binary mask vectors; 
\State update E with one step gradient \textbf{descent} of
\begin{equation*}
\frac{1}{N}\sum_{i=1}^N E({\mathbf{x}^i}) - \frac{1}{N}\sum_{i=1}^N E({G(\mathbf{z}^i, , \mathbf{m}^i)}) - \tilde{H}(p_{data})
\end{equation*}
\EndFor
\end{algorithmic}
\end{algorithm}

\section{Experiments}
\subsection{VGAN Samples} \label{sec:vgan}
\begin{figure}[!thb]
\centering
\includegraphics[scale=0.5]{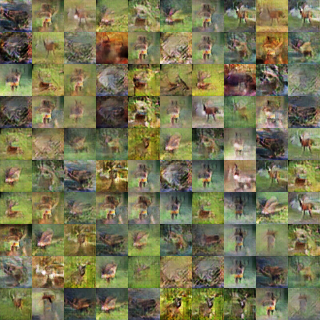}
\hspace{0.1cm}
\includegraphics[scale=0.5]{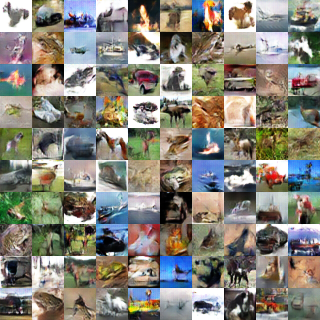}
\includegraphics[scale=0.5]{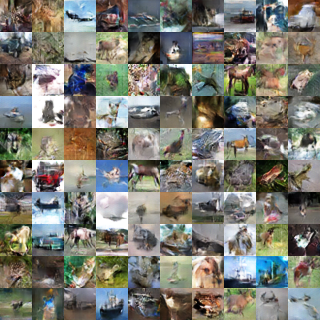}
\hspace{0.1cm}
\includegraphics[scale=0.5]{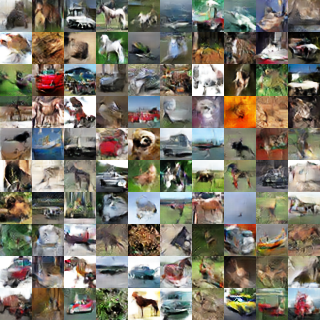}
\caption{Samples from the GAN (left), and generations of VGAN (right), with the same architecture. The first row corresponds to updating the generator one step at each iteration, and the second row corresponds to updating the generator three steps at each iteration.}
\label{fig:examples}
\end{figure}

As a proof of concept, in the first set of experiments, we test the efficacy of the proposed VGAN algorithm as in Algorithm \ref{alg:vgan}. To do this, we train a VGAN on the 50,000 training images of CIFAR-10 with a moderately sized energy (discriminator) and generator. The energy is encoded by a convolutional neural network (CNN) with two convolutional layers, two max pooling layers, and two fully connected layers, where the last fully connected layer is used to compute the energy as defined in Equation \ref{eq:energy} (with K=100). The generator is encoded by a deconvolutional neural network with two consecutive fully connected layers, the latter of which is reshaped and followed by two deconvolutional layers to perform upsampling convolution. Both the energy and the generator use ReLU as nonlinearity, and only the generator is equipped with batch normalization \cite{batchnorm} \footnote{Caution should be taken when attempting to apply batch normalization to the energy (discriminator). An incorrect approach is to apply batch normalization to real data batch and generated batch separately, which essentially makes $E$ different for real and generated data in the energy function $E(\cdot)$.}. Both the energy and the generator are updated with Adadelta \cite{adadelta} using learning rate 0.1. As a direct comparison, we have also trained a GAN with the exact same architecture and training protocol, except that the top layer of the discriminator is replaced with one single sigmoid unit.
We train both VGAN and GAN for 100 epochs, while varying the number of generator updates per iteration from 1 to 3 ($k$ in Algorithm \ref{alg:vgan}). Note that the original GAN paper \cite{gan} proposes to update the discriminator $k$ steps per iteration, which we did the opposite. We show the generated samples from the generator in Figure \ref{fig:examples}. Here the first row corresponds to $k=1$, and the second row corresponds to $k=3$. For each row, on the left are 100 generations from GAN, and on the right are 100 generations from VGAN. We see that for both step numbers, VGAN is able to generate visually appealing images that are difficult to distinguish from samples from the test set. GAN, on the other hand, clearly fails to generate diversified, or realistically looking images when k=1, but works much better when k=3. This can be easily understood from the variational point of view, where a larger step k for generator makes the lower bound tighter, thus producing much stabler models.

\begin{table}
\caption{CIFAR-10 Test error rates of the linear classifiers trained on the second to the top discriminator layer ($\phi(\mathbf{x})$) of GAN and VGAN with generator update steps as 1 and 3. }
\label{tab:clf}
\centering
\begin{tabular}{*4l}
\hline
GAN k=1 & GAN k=3 & VGAN k=1 &VGAN k=3 \\
84.7 & 86.6 & 36.5 & 32.5 \\
\hline
\end{tabular}
\end{table}

In order to further justify the observations, we train two linear classifiers with the second to the top layer fully connected activation from the discriminator of both models (1204 dimensional), for $k=1,3$; the results are shown in Table \ref{tab:clf}. We see that thanks to the bounded multi-modal energy, VGAN is able to benefit from more generator updates. GAN, on the other hand, fails to learn discriminative features, despite the appealing visual quality of generations when k=3. This also verifies our hypothesis discussed in Section \ref{sec:energy}, as the uni-modal nature of GAN discourages it from learning discriminative features at the top layer of its discriminator.

\subsection{Learning with VCD}
In the next set of experiments, we evaluate the variational contrastive divergence of our model. We train our models on three datasets: MNIST, CIFAR-10, and SVHN with 50,000, 40,000, 60,000 training images, respectively. For each dataset, we train a VGAN with variational contrastive divergence, while varying the weight $\rho$ in Equation \ref{eq:ae} from the range \{0, 0.001, 0.01, 0.1, 1\}. Note that in the extreme case when $\rho = 0$, VGAN degrades to training an EBM with negative samples obtained from an autoencoder. In the other extreme case when $\rho = 1$, the transition distribution $p_z(\tilde{\mathbf{x}}|\mathbf{x})$ is not constrained to be centered around $\mathbf{x}$, and is roughly equal to a regular VGAN. We set the dimensionality of $\mathbf{h}, \mathbf{m}, \mathbf{z}$ to be $256$ for MNIST, and $2048$ for CIFAR-10 and SVHN, and use tanh as its nonlinearity (ReLU is used for all other layers except for the top layer of the autoencoder with uses sigmoid). $p_z(\mathbf{z})$ is set to be a uniform distribution drawn from $[-1, 1]$ which matches the magnitudes of $\mathbf{h}$. The training protocol is the same as that described in Section \ref{sec:vgan} except that we use k=1 throughout this set for computational reasons.

\begin{figure}[thb]
\centering
\includegraphics[scale=0.4]{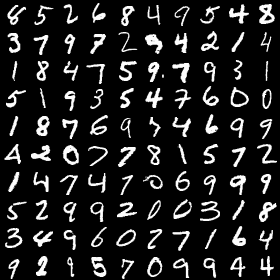}
\includegraphics[scale=0.4]{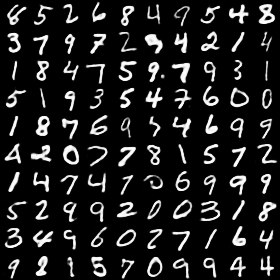}
\includegraphics[scale=0.4]{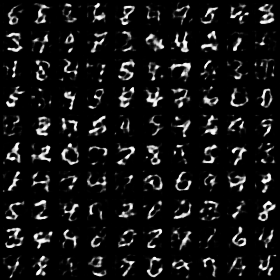}
\includegraphics[scale=0.4]{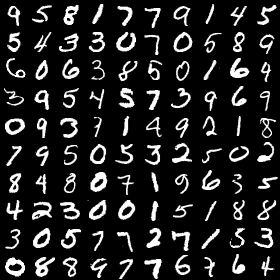}
\includegraphics[scale=0.4]{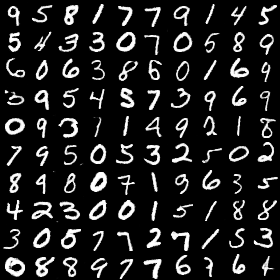}
\includegraphics[scale=0.4]{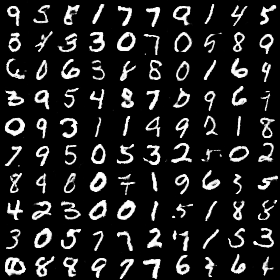}
\includegraphics[scale=0.4]{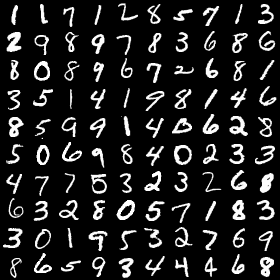}
\includegraphics[scale=0.4]{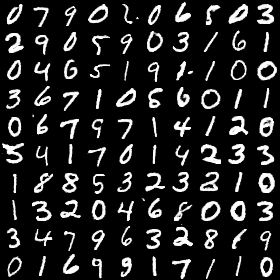}
\includegraphics[scale=0.4]{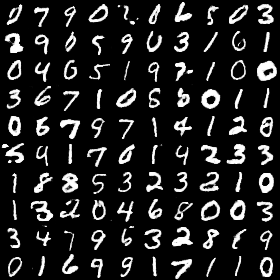}
\caption{Visualization of $\mathbf{x}$, $\mathbf{\bar{x}}$, and $\mathbf{\tilde{x}}$ for $\rho = 0, 0.01, 1$ on MNIST. The first to third row corresponds to $rho = 0, 0.01, 1$, respectively. The first to third column corresponds to samples from the validation set $\mathbf{x}$, reconstructions of the samples $\bar{\mathbf{x}}$, and the generated samples $\tilde{\mathbf{x}}$.} 
\label{fig:mnist}
\end{figure}

\begin{figure}
\centering
\includegraphics[scale=0.4]{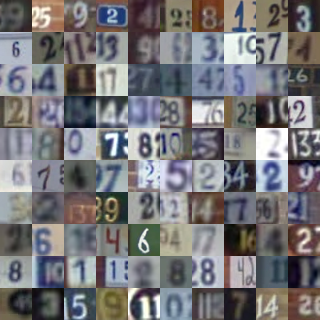}
\includegraphics[scale=0.4]{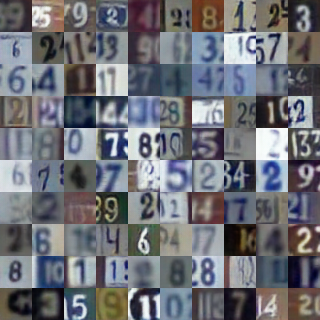}
\includegraphics[scale=0.4]{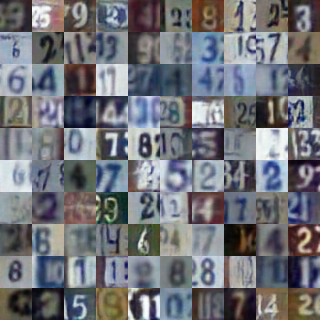}
\includegraphics[scale=0.4]{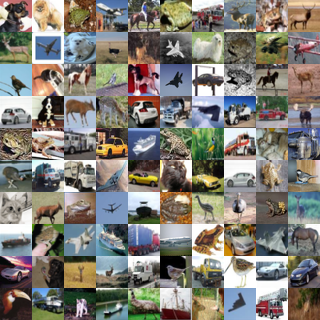}
\includegraphics[scale=0.4]{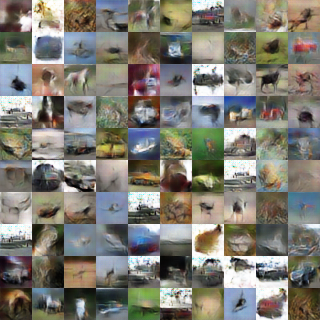}
\includegraphics[scale=0.4]{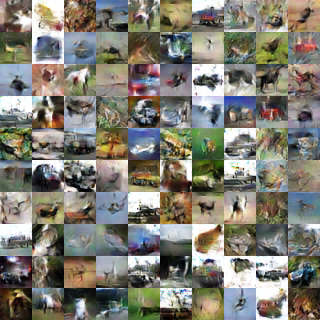}
\caption{Visualization of $\mathbf{x}$, $\mathbf{\bar{x}}$, and $\mathbf{\tilde{x}}$ for $\rho = 0.01$ on SVHN and CIFAR10. The first to third column corresponds to samples from the validation set $\mathbf{x}$, reconstructions of the samples $\bar{\mathbf{x}}$, and the generated samples $\tilde{\mathbf{x}}$.}  
\label{fig:svhn_cifar}
\end{figure}

We first study the effect of varying $\rho$ by looking at the MNIST examples in Figure \ref{fig:mnist}. The first to third row corresponds to $\rho = 0, 0.01, 1$, respectively. The first to third column corresponds to validation samples, reconstructions, and conditional generations, respectively. We see from the first row (which equals to an unregularized autoencoder) that the generator fails to generate realistically looking images. The third row is able to generate realistic images conditioned on a sample, but there is no resemblance between the generation and the sample conditioned on. The second row, on the other hand, is able to both reconstruct the input sample, and also generate realistically looking samples with the transition operator, with notable differences between the input and generation. We have also observed similar trends on SVHN and CIFAR-10 results in Figure \ref{fig:svhn_cifar}, where only $\rho=0.01$ is shown for the space concern.

\begin{figure}
\centering
\includegraphics[scale=0.44]{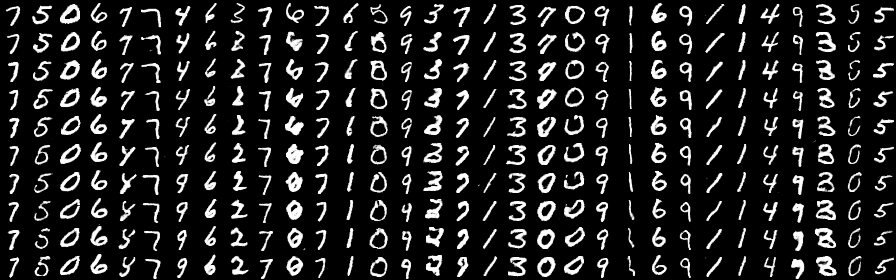}
\includegraphics[scale=0.44]{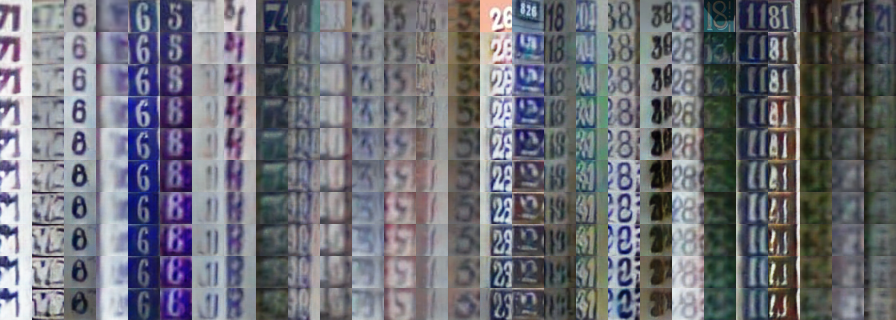}
\caption{Simulating a Markov Chain with $p_z(\mathbf{\tilde{x}}|\mathbf{x})$. We show 30 and 28 images form the validation set for MNIST and SVHN in the first row of each panel, respectively, followed by 9 Gibbs sampling steps. Note the smooth transition of digit types, shapes, and/or colors.}
\label{fig:transition}
\end{figure}

We can also simulate a Markov Chain with the learned transition distribution, and we visualize the results on MNIST and SVHN in Figure \ref{fig:transition}. We see that the learned transition distribution can smoothly vary the style, type, color, and etc. of the digits. Also note that the transitions are not restricted to the Euclidean neighborhood of the samples conditioned on, for example, changing of colors should result in a large distance in the input space, which is our transition operator does not have difficulty exploring. 

Finally, as a quantitative evaluation of the learned transition distribution, we attempt to use the generated conditional samples as data augmentation on MNIST and SVHN \footnote{we are not able to obtain reasonable results on CIFAR-10, as our EMB suffers from noticeable underfitting, identified by the large reconstruction errors in Figure \ref{fig:svhn_cifar}.}. To be concrete, for each dataset we train two additional CNNs enhanced with batch normalization, dropout out and input Gaussian noising. We then minimize the follow loss function:
\begin{equation}
\label{eq:semi}
0.5 * \frac{1}{N}\sum_{i=1}^N\mathcal{L}(\mathbf{x}^i, \mathbf{y}^i) + 0.5 * \frac{1}{N}\sum_{i=1}^N \mathrm{E}_{\mathbf{\tilde{x}^i} \sim p(\mathbf{\tilde{x}} | \mathbf{x}^i)}\mathcal{L}(\mathbf{\tilde{x}}^i, \mathbf{y}^i).
\end{equation}
For each dataset we train on the first 1000 training images, and use the validation set to select the best model; we then report the test error of different configurations. The results are summarized in Table \ref{tab:errors}. We see that on both datasets, with a properly chosen $\rho$ the generator is able to provide good generations to improve learning. On the other hand, with $\rho=0$, which corresponds to sample from an autoencoder, hurts performance. $\rho=1$ completely messes up training as the generated samples are not guaranteed to have the same label as the samples conditioned on. This shows that our transition distribution is able to generate samples that are sufficiently different from training images to boost the performance. Although these numbers are by no means state-of-the-art results, we consider them significant as a proof of concept, because our baseline models are already heavily regularized with dropout and feature noising, which can be considered as data agnostic data augmentation. Also note that there are much space for improvements by leveraging the weights between the two terms in Equation \ref{eq:semi}, tuning the architecture of the energy model, the generator and the classifier model. 

\begin{table}
\centering
\caption{Semisupevised learning error rates by using the learned transition distribution for data augmentation.}
\label{tab:errors}
\begin{tabular}{*8l}
\hline
model & MNIST-1000 & SVHN-1000 \\
No augmentation &2.2 &19\\
VCD ($\rho=0$) &2.9 &26 &\\
VCD ($\rho=0.001$) &2.0 &20\\
VCD ($\rho=0.01$) &1.7 &18\\
VCD ($\rho=0.1$) &1.9 &17 \\
VCD ($\rho=1$) &21 &37\\
\hline
\end{tabular}
\end{table}

\section{Related Work}
There has been a recent surge on improving GANs \cite{dcgan,improvedgan,zhao2016energy,kim2016deep}. \cite{dcgan} proposes a set of techniques to stablize GANs, including using batch normlization, dropping pooling layers, reduced learning rate, and using strided convolutions, but there is little justification of the proposed designs. Our framework, however, directly addresses two most important issues, the energy parametrization and the entropy approximation, and allows the freedom of using the most conventional designs such as pooling and ReLU. \cite{improvedgan} proposes several tricks to enhance the stability. For example, the proposed batch discrimination is in nature similar to our energy design, but with a much higher complexity. \cite{kim2016deep,zhao2016energy} are the two most directly related efforts that connect GANs with EBMs. However, our work is the first to the best of our knowledge to identify the nature of the variational training of EBMs and to provide practical solutions in this view at the same time.

There has also been a long standing interest in terms of EBMs and deep generative models in the machine learning community, such as deep Boltzmann machines and deep belief networks \cite{dbm,dbn}. The contribution of our framework from this aspect is to propose a scalable training method to eliminate the need of MCMC sampling. Variational inference has also been well studied in the literature, but most successfully in dealing with deep directed graphical models such as DBM \citealt{dbm} and variational autoencoder \cite{vae}, where typically variational \textit{upper bounds} are derived for NLL, instead of the \textit{lower bound} in our work. Minimizing the variational lower bound is obviously more difficult to work with, as if the bound is not tight enough, there is no guarantee that the original NLL is minimized. 

Our variational contrastive divergence is also related to GSNs \cite{gsn}, as they both model a transition probability. However, GSNs adopt a transition distribution of form $p(\mathbf{x}|\mathbf{\tilde{x}})$, where $\mathbf{\tilde{x}}$ is produced by adding simple noises to training samples. This essentially limits the space of sampling limited to a Gaussian neighborhood of training examples, which our model does not assume. VCD is also related to the adversarial autoencoder \cite{adae} as they both include an autoencoder module, but with fundamental differences: the use of autoencoder in our work is part of and to improve the EBM/GAN, while \cite{adae} on the other hand, requires another GAN besides the autoencoder.

\section{Conclusion}
We have proposed VGANs, a family of methodologies to train deep EBMs with an auxiliary variational distribution. We have drawn connection between deep EBMs and GANs, and propose practical solutions to stabilizing training. We show that our proposed bounded multi-modal energy combined with variational contrastive divergence works well on generating realistically looking images, and recovering the data manifold by simulating a Markov Chain. We have also attempted to utilize the learned transition distributions to perform data augmentation in the context of semisupervised learning, and show consistent improvements.
\bibliography{iclr2017_conference}
\bibliographystyle{iclr2017_conference}

\end{document}